# Learning the Experts for Online Sequence Prediction

Elad Eban                                                     ELADE@CS.HUJI.AC.IL
Aharon Birnbaum                                           AHAROB01@CS.HUJI.AC.IL
Shai Shalev-Shwartz                                           SHAIS@CS.HUJI.AC.IL
Amir Globerson                                              GAMIR@CS.HUJI.AC.IL

Selim and Rachel Benin School of Computer Science and Engineering. The Hebrew University of Jerusalem.

## Abstract

Online sequence prediction is the problem of predicting the next element of a sequence given previous elements. This problem has been extensively studied in the context of individual sequence prediction, where no prior assumptions are made on the origin of the sequence. Individual sequence prediction algorithms work quite well for long sequences, where the algorithm has enough time to learn the temporal structure of the sequence. However, they might give poor predictions for short sequences. A possible remedy is to rely on the general model of prediction with expert advice, where the learner has access to a set of $r$ experts, each of which makes its own predictions on the sequence. It is well known that it is possible to predict almost as well as the best expert if the sequence length is order of $\log(r)$. But, without firm prior knowledge on the problem, it is not clear how to choose a small set of *good* experts. In this paper we describe and analyze a new algorithm that learns a good set of experts using a training set of previously observed sequences. We demonstrate the merits of our approach by applying it on the task of click prediction on the web.

## 1. Introduction

Sequence prediction is a key task in machine learning and statistics. It involves predicting the next element in a sequence given the previous elements. Typical applications include stock market prediction, click prediction in web browsing and consumption predic-tion in smart grids. Although the sequence prediction problem has been well studied, current solutions either work for long sequences or require strong prior knowledge. In this work we provide a method that uses training data to learn how to predict a novel sequence. As we shall show, we use the training sequences to obtain the prior knowledge needed for predicting novel sequences.

Sequence prediction is most naturally cast as an online prediction problem (Cesa-Bianchi and Lugosi, 2006), where at every step we predict the next element, and then receive the true value of the element while suffering a loss if we made a prediction error. We are then allowed to improve the model, and predict the next step. The online formulation is natural in most applications since the new element's true value unfolds in real time and we are interested in minimizing the prediction loss of this process.

One classical approach to the problem is the so called *universal sequence prediction* class of methods (Feder et al., 1992; Hutter, 2006). Such methods guarantee that asymptotically (with sequence size) the model will achieve optimal prediction error. However, the price we pay for universality is that good performance will be reached only after seeing long sequences. Intuitively, the reason for this is that no prior knowledge about the sequence is used, so it may take a while until we have a good model of it.

An alternative approach which does introduce prior knowledge is predicting with expert advice (Littlestone and Warmuth, 1994; Vovk, 1990). Here one has a set of $r$ experts, where each expert is a sequence predictor. The Weighted-Majority algorithm (Littlestone and Warmuth, 1994) uses such experts to do online prediction, and is guaranteed to perform almost as well as the best expert. More formally, for any sequence of length $T$, the average number of prediction mistakes of Weighted-Majority ($WM$) is bounded above by the average number of prediction mistakes made by the best expert plus $\sqrt{\log(r)/T}$. Thus, the *WM* algorithm will perform well when for every sequence there exists an expert that performs well on it and when the sequence is long enough.

Given the above, it's clear that learning from experts will work when the experts *fit* the sequences we want to predict. Thus a key question, which we address here, is how to choose a good set of experts. We propose to learn these experts from a set of training sequences. In the spirit of empirical risk minimization we shall seek a set of experts that perform well on our training set. This is a highly non-trivial task in our case due to several reasons. First, the performance of a pool of experts is measured by the performance of an online algorithm whose parameters are the experts, and it's not clear how to optimize this function. We shall see that the hindsight loss is a simpler function to optimize and results in comparable theoretical guarantees. Second, we would like our experts to use arbitrarily long histories in making predictions, but do so without over fitting. We shall show that this can be done by using a variant of context trees (a.k.a. prediction suffix trees). Finally, it's not clear what generalization guarantees, if any, can be expected from such a scheme. We perform a detailed generalization analysis, providing theoretical bounds on the sample complexity of learning a good set of experts.

Our learning task is thus as follows: given a training set of sequences, learn a set of experts that will work well for online sequence prediction. In a sense, this can be viewed as a collaborative version of sequence prediction. We provide an objective that corresponds to this discriminative setting and analyze the generalization error of its minimizer. Our theoretical analysis provides generalization bounds that show no over fitting for longer histories, and quantify the advantage of learning in the collaborative setting.

We apply our model to synthetic and real-world problems and show that it outperforms methods which do not use temporal and collaborative approaches.

## 2. Problem Formulation

Let $\Sigma$ be a finite alphabet. A sequence of symbols is a member of $\Sigma^*$ and is denoted by $\mathbf{x} = (x_1, \ldots, x_T)$. Online sequence prediction takes place in consecutive rounds. On round $t$, the forecaster observes the prefix $\mathbf{x}_{1:t-1} = (x_1, \ldots, x_{t-1})$ and predicts $\hat{x}_t \in \Sigma$. Then, the next symbol, $x_t$, is revealed and the forecaster pays $\mathbb{1}_{[x_t \neq \hat{x}_t]}$. That is, it pays 1 if $x_t \neq \hat{x}_t$ and 0 otherwise.

An "expert" for sequence prediction is a function $f : \Sigma^* \to \Sigma$. Such an expert can be used for predicting the $t$'th symbol by setting $\hat{x}_t = f(x_{1:t-1})$.

Given a set of $r$ such experts, the Weighted-Majority (WM) algorithm (Littlestone and Warmuth, 1994) (see pseudocode below), can be used for predicting almost as well as the best expert.

A performance guarantee for WM is provided in the following theorem (Littlestone and Warmuth, 1994).

---

**Weighted Majority (WM)**

**parameter:** $\eta > 0$
**initialize:** $\boldsymbol{w}_1 = (1/r, \ldots, 1/r)$
**for** $t = 1, 2, \ldots, T$
  choose $i \sim \boldsymbol{w}_t$ at random
  predict $\hat{x}_t = f_i(\mathbf{x}_{1:t-1})$
  **update rule** $\forall i, \ w_{t+1}[i] \propto w_t[i] e^{-\eta \mathbb{1}_{[f_i(\mathbf{x}_{1:t-1}) \neq x_t]}}$

---

**Theorem 1** *The Weighted Majority algorithm (with $\eta = \sqrt{\log(r)/T}$) obtains the following regret bound:*

$$\frac{1}{T} \sum_{t=1}^{T} \mathbb{P}[\hat{x}_t \neq x_t] \leq \min_i \frac{1}{T} \sum_{t=1}^{T} \mathbb{1}_{[f_i(\mathbf{x}_{1:t-1}) \neq x_t]} + \sqrt{\frac{4\log(r)}{T}}.$$

It follows that we can predict the sequence reasonably well if two conditions hold:

1. $\log(r)$ is sufficiently small compared to $T$.

2. At least one of the experts makes a small number of mistakes on the sequence.

Therefore, when choosing a set of experts we face the classical bias-complexity tradeoff: On one hand we want $r$ to be small enough so that the regret term $\sqrt{\log(r)/T}$ will be small. On the other hand, different experts will work well on different sequences, and since we do not know the type of sequence we are going to get, we would like to increase $r$ so that the set of experts will be rich enough to explain many types of sequences.

In this paper we propose to learn a good set of experts based on a sample of sequences. Formally, let $\mathcal{H}$ be a hypothesis class of experts. It is convenient to allow experts to output predictions from a set $Y$, where we have some way to convert an element from $Y$ into a prediction in $\Sigma$. For example, we can use $Y = \mathbb{R}^{|\Sigma|}$, where we interpret the prediction $\boldsymbol{y} \in Y$ as a score for each of the symbols in $\Sigma$. The mapping from a score vector in $Y$ to an actual prediction in $\Sigma$ is via $\arg\max_{\sigma \in \Sigma} y_\sigma$. Therefore, each $f \in \mathcal{H}$ is a function from $\Sigma^*$ to $Y$. The loss of a prediction $f(\mathbf{x}_{1:t-1})$ is measured by a loss function $\ell : Y \times \Sigma \to \mathbb{R}$. The loss function can be the 0-1 loss $\mathbb{1}_{[x_t \neq \hat{x}_t]}$. Later, we use other loss measures that are convex surrogates of the zero-one loss.

The problem that we consider in this paper can be formalized as follows: We are given a sample of sequences, $S = (\mathbf{x}^{(1)}, \ldots, \mathbf{x}^{(m)})$, where each $\mathbf{x}^{(i)}$ is assumed to be sampled i.i.d. from an unknown distribution $\mathcal{D}$ over $\Sigma^*$. Our goal is to use $S$ for learning a set of experts, $F \subset \mathcal{H}$, of size $|F| \leq r$, where $r$ is a parameter of the learning problem (which should depend on the typical size of $T$). We wish to learn $F$ such that when running WM on a new sequence with the set $F$ it will have a small number of mistakes.

Given an expert $f$ and a sequence $\mathbf{x}$, we denote by $L(f, \mathbf{x})$ the average loss of $f$ on $\mathbf{x}$, specifically:

$$L(f, \mathbf{x}) = \frac{1}{T} \sum_{t=1}^{T} \ell(f(\mathbf{x}_{1:t-1}), x_t) .$$

Given a set of experts, $F \subset \mathcal{H}$, we denote by $\mathrm{WM}(F, \mathbf{x})$ the averaged loss of applying the WM algorithm on the sequence $\mathbf{x}$ with the set of experts $F$. Therefore, our ultimate goal is to learn a set of experts $F$ which (approximately) minimizes

$$\mathbb{E}_{\mathbf{x} \sim \mathcal{D}}[\mathrm{WM}(F, \mathbf{x})] .$$

Before we describe how we learn $F$, let us first consider two extreme situations. First, for $r = 1$, i.e. $F = \{f\}$, then $\mathrm{WM}(F, \mathbf{x}) = L(f, \mathbf{x})$. That is, at prediction time, we simply follow the predictions of the single expert $f$. This is exactly the standard traditional setting of statistical batch learning, where we would like to learn a model $f$ from a hypothesis class $\mathcal{H}$ whose expected loss over a randomly chosen example (in our case $\mathbf{x} \sim \mathcal{D}$) is as small as possible. The problem with this approach is that it might be the case that the sequences are of different types, where no single expert from $\mathcal{H}$ is able to accurately predict all of the sequences. On the other extreme, if we set $r = \infty$, i.e. $F = \mathcal{H}$, then we revert to the problem of online learning with a hypothesis class $\mathcal{H}$. The problem with this approach is that if $\mathcal{H}$ is "complex",[1] then the sequence length required in order to guarantee good performance of the online learning might be very large.

## 3. The Learning Algorithm

A straightforward approach for learning $F$ when $r > 1$ is to follow the empirical risk minimization (ERM)

principle, namely, to solve the optimization problem

$$\min_{F \subset \mathcal{H} : |F| = r} \frac{1}{m} \sum_{i=1}^{m} \mathrm{WM}(F, \mathbf{x}^{(i)}) .$$

This problem might be difficult to optimize since the objective function involves the activation of an algorithm and does not have a simple mathematical formulation. To overcome this difficulty, we show how a simpler objective may be used. In light of Theorem 1 (generalized to convex surrogate losses) we know that for any sequence $\mathbf{x}$,

$$\mathrm{WM}(F, \mathbf{x}) \leq \min_{f \in F} L(f, \mathbf{x}) + \sqrt{\frac{4 \log(|F|)}{T}} . \quad (1)$$

Let us slightly overload notation and denote

$$L(F, \mathbf{x}) = \min_{f \in F} L(f, \mathbf{x}) .$$

Thus $L(F, \mathbf{x})$ is the hindsight loss when learning the sequence $\mathbf{x}$ with experts $F$. Taking expectation of both sides of Eq. 1 we obtain that

$$\mathbb{E}_{\mathbf{x} \sim \mathcal{D}}[\mathrm{WM}(F, \mathbf{x})] \leq \mathbb{E}_{\mathbf{x} \sim \mathcal{D}}[L(F, \mathbf{x})] + \mathbb{E}_{\mathbf{x} \sim \mathcal{D}} \sqrt{\frac{4 \log(|F|)}{T}} . \quad (2)$$

The second summand only depends on $F$ via its size. Therefore, for a fixed size of $F$, we can follow a standard bound minimization approach and aim at minimizing $\mathbb{E}_{\mathbf{x} \sim \mathcal{D}}[L(F, \mathbf{x})]$ instead of $\mathbb{E}_{\mathbf{x} \sim \mathcal{D}}[\mathrm{WM}(F, \mathbf{x})]$. In other words, we minimize the hindsight loss instead of the online loss. An ERM approach to this minimization yields the following minimization problem on the training set of sequences:

$$\min_{F \subset \mathcal{H} : |F| = r} \frac{1}{m} \sum_{i=1}^{m} L(F, \mathbf{x}^{(i)}) . \quad (3)$$

By definition of $L(F, \mathbf{x})$, this can be written equivalently as

$$\min_{f_1, \ldots, f_r \in \mathcal{H}} \frac{1}{m} \sum_{i=1}^{m} \min_{\boldsymbol{w} \in \Delta^r} \sum_{j=1}^{r} w_j L(f_j, \mathbf{x}^{(i)}) \quad (4)$$

where $\Delta^r = \{\boldsymbol{w} \in \mathbb{R}^r : \boldsymbol{w} \geq 0, \ \|\boldsymbol{w}\|_1 = 1\}$.

Assuming that $\mathcal{H}$ can be encoded as a convex set and $L(f, \mathbf{x})$ is a convex function,[2] we obtain that the objective of Eq. 4 is convex in $f_1, \ldots, f_r$ and $\boldsymbol{w}^{(1)}, \ldots, \boldsymbol{w}^{(r)}$ individually but not jointly. This suggests an alternating optimization scheme where one alternates between optimizing over $\boldsymbol{w}$'s and over $f$'s. This scheme is especially attractive since minimizing over $\boldsymbol{w}$ for fixed $F$ is straightforward: for each sequence $\mathbf{x}^{(i)}$ find the best expert and set $\boldsymbol{w}^{(i)}$ to 1 for that expert and 0 otherwise. Optimizing $f_i$ for fixed $\boldsymbol{w}$ can be done via gradient descent when using a smooth loss as we do here (see Sections 3.1 and 3.2).

---

[1] As measured, for example, by its Littlestone dimension (Ben-David et al., 2009).

[2] This will be the case for the class $\mathcal{H}$ and loss function we use in Section 3.1.

### 3.1. The class of bounded norm context trees

Thus far we have given a general scheme and have not described the particular set of experts we will use. In what follows we specify those. Any function $f : \Sigma^* \to \Sigma$ can be described using a *multiclass context tree*. For our experts, we will be using a generalization of multiclass context trees following Dekel et al. (2010), described below.

To simplify notation, denote $\Sigma = [k] = \{1, \ldots, k\}$. A multiclass context-tree is a $k$-ary rooted tree, where each node of the tree is associated with a vector $\boldsymbol{z} \in \mathbb{R}^k$. The prediction of the tree on a sequence $\mathbf{x}_{1:t-1}$ is determined as follows. We initially start with the vector $\boldsymbol{z} = \mathbf{0} \in \mathbb{R}^k$, and set the current node to be the root of the tree. We then add to $\boldsymbol{z}$ the vector associated with the current node and traverse to its $x_{t-1}$ child, which becomes the current node. We add again the vector associated with the current node and traverse to its $x_{t-2}$ child. This process is repeated until we arrive either to $x_1$ or to a leaf of the tree. The final value of $\boldsymbol{z}$ gives a score value to each of the elements in $\Sigma$, and the actual prediction is $\arg\max_i z_i$.

It is convenient to represent a context tree as a matrix with $k$ rows as follows. Let us order the nodes of a full $k$-ary tree in a breadth first manner. For simplicity, we restrict ourselves to trees of bounded depth (which can be very large, so this is not a serious limitation). To represent a context tree as a matrix, we set column $i$ of the matrix to be the vector associated with the $i$'th node of the tree (where if the node does not exist in the tree we simply set the column to be the all zeros vector). Similarly, we can map a sequence $\mathbf{x}_{1:t-1}$ to a vector $\psi(\mathbf{x}_{1:t-1}) \in \{0, 1\}^{|\Sigma^*|}$ as follows. Suppose that we traverse from the root of a full $k$-ary tree according to the symbols $x_{t-1}, x_{t-2}, \ldots, x_1$, as we described before. Then, we set all the coordinates of $\psi(\mathbf{x}_{1:t-1})$ corresponding to nodes we visited in this path to be 1, and set all the rest of the coordinates to be zero.

It is easy to verify that the vector $\boldsymbol{z}$ constructed by a context tree for the history $\mathbf{x}_{1:t-1}$ is $\mathbf{U} \psi(\mathbf{x}_{1:t-1})$, where $\mathbf{U}$ is the matrix describing the context tree (the size of $U$ is thus $|\Sigma| \times |\Sigma^*|$ and the columns correspond to the vectors $\boldsymbol{z}$ at each node).

As mentioned before, any function $f : \Sigma^* \to \Sigma$ can be described by a context tree (as long as we allow its depth to be large enough). Therefore, without additional constraints, learning the class of all context trees from a finite sample will lead to over-fitting. To overcome this, one can constrain the depth of the tree. Alternatively, we can allow any depth but carefully discount long histories as described next.

Following (Dekel et al., 2010), we aim to balance between long histories (can be very informative but are rare in the data hence are hard to learn) and short histories (less informative but easier to learn). This can be done by defining a norm over matrices corresponding to context trees, where longer histories are penalized more. Formally, for each column $i$ of a context tree matrix $\mathbf{U}$, let $d(i)$ be the depth of its corresponding node in the tree. Let $a_1 \geq a_2 \geq \ldots$ be a sequence such that $\sum_{i=1}^{\infty} a_i \leq 1$.[3] Then, we define a norm of vectors to be such that

$$\|\mathbf{u}\|^2 = \sum_i a_{d(i)} u_i^2 \ , \tag{5}$$

and a norm over matrices to be $\|\mathbf{U}\|^2 = \sum_j \|\mathbf{U}_j\|^2$, where $\mathbf{U}_j$ is the $j$'th row of $\mathbf{U}$. Put another way, the squared norm of $\mathbf{U}$ is a weighted sum of the squared Euclidean norms of columns of $\mathbf{U}$, where the weight of column $i$ is $a_{d(i)}$. Thus, we assign a higher penalty to columns corresponding to deep nodes of the trees.

Consequently, we define the hypothesis class of bounded norm context trees to be

$$\mathcal{H}_B = \{\mathbf{U} : \|\mathbf{U}\| \leq B\} \ . \tag{6}$$

Finally, we also need to define scale sensitive loss functions. A common choice is the multiclass log-loss:

$$\ell(\boldsymbol{z}, y) = \log \left( \sum_{y' \in \Sigma} \exp \left( \mathbb{1}_{[y' \neq y]} - z_y + z_{y'} \right) \right) .$$

This loss function has the advantages of being a convex surrogate loss for the zero-one loss.

### 3.2. The LEX algorithm

We are now ready to describe our algorithm, which we call **LEX** (for Learning Experts). Our goal is to minimize the loss in Eq. 4 with respect to the vectors $\boldsymbol{w}_i$ and the parameters of the experts. As described in Section 3.1, we parameterize each expert by a context tree matrix $\mathbf{U} \in \mathbb{R}^k \times |\Sigma^*|$. As mentioned earlier, we can minimize Eq. 4 via alternating optimization where minimizing over $\boldsymbol{w}$ can be done in closed form

---

[3]Here we take $a_i = i^{-2}$.

and minimizing over $\mathbf{U}$ can be done with gradient descent. Calculating the gradient w.r.t. $\mathbf{U}$ is easy for the log loss. In our implementation we use stochastic gradient descent, where an update is performed after each training sequence is processed.

## 4. Analysis

Define the generalization loss for the set of experts $F$:

$$L_{\mathcal{D}}(F) = \underset{\mathbf{x} \sim \mathcal{D}}{\mathbb{E}} L(F, \mathbf{x}) \ .$$

In light of Eq. 2, in order to bound $\mathbb{E}_{\mathbf{x}}[\mathrm{WM}(F, \mathbf{x})]$ it suffices to bound $L_{\mathcal{D}}(F)$. In this section we derive bounds on $L_{\mathcal{D}}(F)$. Our bounds depend on the following measures: the number of experts $|F|$, a complexity measure of the hypothesis class $\mathcal{H}$, the number of training examples, and the training loss:

$$L_S(F) = \frac{1}{|S|} \sum_{\mathbf{x} \in S} L(F, \mathbf{x}) \ .$$

We first define a complexity measure for a hypothesis class $\mathcal{H}$ with respect to a loss function $\ell$.

**Definition 1** *Let $\mathcal{H}$ be a class of functions from $Z$ to $Q$, let $Y$ be a target set, and let $\ell : Q \times Y \to \mathbb{R}$ be a loss function. We say that the complexity of $\mathcal{H}$ is $C(\mathcal{H})$ if for any sequence $(z_1, y_1), \ldots, (z_q, y_q) \in (Z \times Q)^q$ and for any $\epsilon > 0$, there exists $\mathcal{H}' \subset \mathcal{H}$ of size $|\mathcal{H}'| \leq (2q)^{C(H)/\epsilon^2}$, such that for all $h \in \mathcal{H}$ exists $h' \in \mathcal{H}'$ that satisfies*

$$\forall i \in [q], \quad |\ell(h(z_i), y_i) - \ell(h'(z_i), y_i)| \leq \epsilon \ .$$

The reader familiar with covering number bounds can easily recognize $C(\mathcal{H})$ as determining the size of a cover of $\mathcal{H}$. It is also easy to verify that if $\mathcal{H}$ is a class of binary classifiers then $C(\mathcal{H})$ is upper bounded by the VC dimension of $\mathcal{H}$ (this follows directly from Sauer's lemma). We will later show that the class of bounded norm context trees has a bounded $C(\mathcal{H})$ as well.

**Theorem 2** *Let $\mathcal{D}$ be a probability over $\Sigma^*$ such that there exists some constant $T$ with $\mathbb{P}[\mathrm{len}(x) \leq T] = 1$. Assume also that for all $\mathbf{x}$ and $F \subset \mathcal{H}$ we have $L(F, \mathbf{x}) \in [0, \sqrt{C(\mathcal{H})}]$. Then, with probability of at least $1 - \delta$ over $S \sim \mathcal{D}^m$, for all $F \subset \mathcal{H}$, with $|F| = r$, we have*

$$L_{\mathcal{D}}(F) \leq L_S(F) + \tilde{O}\left(\sqrt{\frac{r\, C(\mathcal{H})}{m}}\right) \ .$$

The above theorem tells us that if $\mathcal{H}$ is of bounded complexity, then the number of samples required to have $L_{\mathcal{D}}(F) \leq L_S(F) + \epsilon$ is order of $r\, C(\mathcal{H})/\epsilon^2$. In particular, the sample complexity of learning a set of $r$ experts is $r$ times larger than the sample complexity of learning a single expert.

The proof of the theorem is given in the long version of this article. The main ideas of the proof are as follows. First, we construct a cover for the loss class $\{\mathbf{x} \mapsto L(F, \mathbf{x}) : F \in \mathcal{H}, |F| = r\}$. Then, we bound the Rademacher complexity of this class using a generalization of Dudley's chaining technique, which is similar to a technique recently proposed in Srebro et al. (2010).

Next, we turn our attention to the specific class of context trees with bounded norm. The following lemma bounds its complexity.

**Lemma 1** *Let $\mathcal{H}_B$ be the class of multiclass context trees which maps $\Sigma^*$ into $\mathbb{R}^{|\Sigma|}$ as defined in Section 3.1. Let $\ell : \mathbb{R}^{|\Sigma|} \times \Sigma \to \mathbb{R}$ be a loss function such that*

$$\forall l \in \Sigma, \ u, v \in \mathbb{R}^{|\Sigma|}, \quad |\ell(u, l) - \ell(v, l)| \leq \|u - v\|_{\infty} \ .$$

*Then: $C(\mathcal{H}_B) \leq O(B^2 \log(k))$.*

The proof of the lemma is given in the long version of this article. The main idea is a nice trick showing how to bound the cover of a linear class based on known bounds on the convergence rate of sub-gradient mirror descent algorithms (e.g., see Nemirovski and Yudin, 1978). This is similar to a method due to Zhang (2002), although our bound is slightly better.

The multiclass log-loss function satisfies the conditions of the above lemma, hence:

**Corollary 1** *Let $\mathcal{H}_B$ be the class of multiclass context trees and let $\ell$ be the multiclass log-loss. Let $\mathcal{D}$ be a probability over $\Sigma^*$ such that there exists some constant $T$ with $\mathbb{P}[\mathrm{len}(x) \leq T] = 1$. Then, with probability of at least $1 - \delta$ over $S \sim \mathcal{D}^m$, for all $F \subset \mathcal{H}_B$, with $|F| = r$, we have*

$$L_{\mathcal{D}}(F) \leq L_S(F) + \tilde{O}\left(\sqrt{\frac{r\, B^2}{m}}\right) \ .$$

In summary, if we manage to find a set $F \subset \mathcal{H}_B$ of size $r$ that achieves a small hindsight training loss, then it will also achieve a small hindsight generalization loss. Combining this with Eq. 2 yields

$$\mathbb{E}[\mathrm{WM}(F, \mathbf{x})] \leq L_S(F) + \tilde{O}\left(\sqrt{\frac{r\, B^2}{m}}\right) + \mathbb{E}\sqrt{\frac{4\log(r)}{T}} \ .$$

Therefore, the performance of the Weighted-Majority algorithm is upper bounded by three terms: The training loss of $F$ (which can decrease when increasing $r$), the estimation error term (which increases with $r$), and the online regret term (which also increases with $r$).

## 5. Related Work

The problem of sequence prediction has a fairly long history and has received much attention from game theorists (Robbins, 1951; Blackwell, 1956; Hannan, 1957), information theorists (Cover and Hart, 1967; Cover and Shenhar, 1977; Feder et al., 1992; Willems et al., 1995), and machine learning researchers (Helmbold and Schapire, 1997; Pereira and Singer, 1999; Cesa-Bianchi and Lugosi, 2006; Dekel et al., 2010). One of the most useful tools is context trees, which store informative histories and the probability of the next symbol given these. However, all of these works consider predicting a sequence from a single source. Indeed, our work extends these single sequence predictions to the collaborative setting where we model different sequences, but constrain the predictors to share some common structure (i.e., the experts used in prediction).

Another related line of work is multitask prediction (e.g., see Ando and Zhang, 2005; Abernethy et al., 2007), in which one considers several different multiclass prediction problems and seeks a common feature space for those. This setting is different from ours in several ways. First, in the multitask setting one receives a set of training instances from each task, where it is known which sample belongs to each class. In our case, we receive only a set of individual sequences. Furthermore, in the multitask setting, the test data comes from one of the known tasks, whereas we again receive a novel sequence from an unknown source.

A more recent approach to sequence modeling is the "sequence memoizer", which is based on nonparametric Bayesian models (Wood et al., 2009). So far these have been applied to a single type model (e.g., language modeling), and not for multiple distinct models as we have here. It is conceivable that a fully Bayesian model for collaborative sequence prediction can be built using these models, and it would be interesting to contrast it with our approach.

Another possible approach to the problem is to use probabilistic latent variable models (Hofmann, 1999) or their discriminative counterparts (Felzenszwalb et al., 2008; Yu and Joachims, 2009). Here each sequence will be mapped to a latent variable corresponding to the best expert. Next, given the class and the previous history, a probabilistic suffix tree will be used to generate the next action. However, such a model will not handle long histories appropriately and is likely to result in overfitting (as our empirical results also show). While it may be possible to add history discounting to such a model, it will be considerably more complex than what we suggest here.

In our formulation, the state space $\Sigma$ is unstructured. There are cases of interest, where $\Sigma$ has structure. For example, it may correspond to the items in an online shopping basket. Prediction in such a setting was recently addressed in Rendle et al. (2010). Unlike in our case, they have access to multiple training sequences from particular users, and prediction is done on these users. Furthermore, the temporal model itself is only first order and thus very different from ours. Note that we can easily extend our approach to structured state spaces by using structured prediction instead of multiclass as we do here.

## 6. Experiments

In what follows, we evaluate the performance of the **LEX** algorithm (see Section 3.2) on two datasets: synthetic and real-world. We compare it to the baselines described below.

### 6.1. Baselines Models

We consider three different baselines models. The first is our **LEX** algorithm with $r = 1$ (we denote this baseline by 1-**LEX**), which is in fact a batch trained PST (where training uses the log loss). In this approach all training sequences are modeled via a single PST corresponding to one expert. It thus does not directly model multiple temporal behaviors of the sequences in the data.

Our second baseline is an *online* PST model which is evaluated on each test sequence individually. Training is done using the algorithm in (Dekel et al., 2010). Being an online algorithm, it does not use the training data. However, given long enough sequences it will be able to model any deterministic temporal behavior optimally. In other words, this algorithm has the benefit of adaptation but its performance crucially depends on the length of the sequence. We denote this baseline by **Online PST**.

Finally, we consider a generative latent variable model (denoted by **LMM**) which is a mixture of Markov chains. An order $d$ Markov chain is a basic yet powerful tool for modeling sequences. In **LMM** we generalize Markov chains by allowing each sequence to be generated by one of $r$ regular Markov models. We think of these $r$ models as different *chain types* similarly to the $r$ experts of **LEX**. Specifically, for a sequence $x_1, \ldots x_{t-1}$ the $r$-**LMM** model of order $d$ is defined by:

$\Pr(x_t|x_{1:t-1}) \stackrel{\text{def}}{=} \sum_{q=1}^{r} \mathbb{P}(x_t|x_{t-d:t-1}, z = q) \mathbb{P}(z = q)$
Where $z$ is the latent (unobserved) variable which "assigns" a *chain type* to a sequence. Note that the standard Markov chain is simply a 1-**LMM**. We learn the parameters of a **LMM** from training data using *EM*. The (online) prediction using this model is done by the maximum a-posteriori assignment at each point in time. Since **LMM** does not discount long histories, it is not expected to perform well when $d$ is large and not enough training data is available. Parameters for all algorithms (i.e., $r$ and $d$) were tuned using cross validation.

## 6.2. Synthetic Data

We begin by considering sequences that follow one of two temporal patterns. The sequences are generated as follows: First randomly select $j \in \{1, 2\}$ then draw $T$ samples according to the (independent) distribution:
$\Pr(x_t = x) = \begin{cases} 2^{-1} & \text{if } x = j \\ (2(|\Sigma| - 1))^{-1} & \text{otherwise} \end{cases}$. We used $|\Sigma| = 200$ and generated a set of $m = 1000$ sequences, each of length $T = 250$ (these parameters where selected to resemble the browsing data characteristics). We note that by construction, the maximal possible generalization accuracy on this data is 0.5. We evaluate the accuracy of online prediction on 400 test sequences.

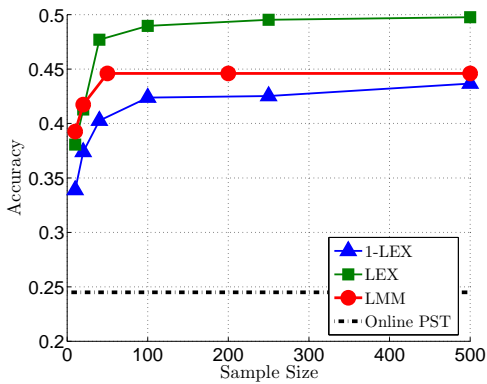

*Figure 1.* Test accuracy of online prediction on the synthetic data. See Section 6.2. The four algorithms are described in Section 6.1.

In Fig. 1 we show the accuracy (on test data) of **LEX** and the three baselines. We notice that **LEX** approaches 0.5 accuracy using about 50 sequences, 1-**LEX** and **LMM** require substantially more samples in order to approach this performance (over 500 sequences for 0.45 accuracy). In other words, in agreement with our theoretical analysis, the sample complexity of **LEX** is smaller than both 1-**LEX** and **LMM**. The accuracy of online **PST** is much lower

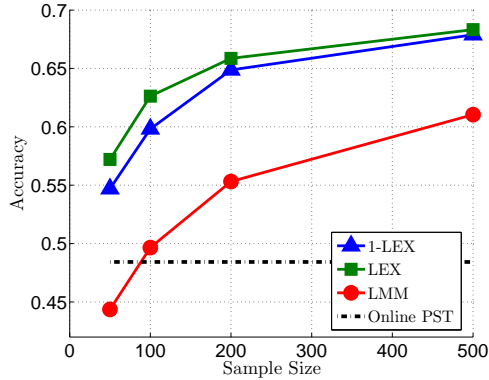

*Figure 2.* Test accuracy of online prediction on the click prediction task. See Section 6.3. The four algorithms are described in Section 6.1.

due to the conservative training of this algorithm.

## 6.3. Click Prediction Data

Here we consider a challenging task of predicting the browsing pattern of web users. Specifically, we use browsing logs for users in an intra-net site. For each session the sequence of *urls* visited by every user was recorded by the web server. The dataset contains 2000 such sequences of length 70-150. The domain of the prediction problem, is of distinct *urls* and its magnitude is $|\Sigma| = 189$. The data was split into train, validation and test sets, the sizes of the training sets vary, while the validation and test set sizes were fixed at 200 and 800 sequences respectively. We applied the three baseline models, and compared their performance to **LEX**. In this experiment the $r$ experts learned by **LEX** were combined with an additional expert obtained from training a 1-**LEX** algorithm, resulting in a pool of $r+1$ learned experts. This addition smoothes performance on short sequences where the WM algorithm might not have enough time to decide which of the $r$ experts to follow.

Results are shown in Fig. 2. It can be seen that **LEX** outperforms the other methods. When considering the difference in accuracy between **LEX** and 1-**LEX** we notice that the added accuracy from multiple experts shrinks as training size increases. This trend agrees with theory, since as more data is available to 1-**LEX**, it can use longer histories and eventually will be able to model any temporal behavior. However, as we show in the synthetic experiments, the gap for small data sizes can be considerable.

## 7. Discussion

We have described and analyzed a method for learning the experts for online sequence prediction. In particular, we specified it to the class of prediction suffix trees. Thus, our experts can capture dependencies on arbitrarily long histories. This is achieved by mapping context trees into a vector space and designing a norm on this space which discounts long histories. As our generalization results show, the complexity of the model is not penalized by the maximal possible length of histories (dimensionality of the matrix $\mathbf{U}$) but rather by the effective needed context based history (captured by the norm of $\mathbf{U}$). Our empirical results show that temporal user specific structure can indeed be used to improve prediction accuracy.

The proposed approach can be extended in several ways. First, we can consider different prediction goals: instead of predicting the next symbol in the sequence, corresponding to the next URL, we can have a binary classifier that returns one if a user is likely to take a given action and zero otherwise. Alternatively, we can consider a ranking task where we want to sort actions according to their interest to the user. To use such objectives we will just need to replace our multiclass log loss with the corresponding loss.

Finally, we note that our model can be applied to a wide array of practical problems. Some examples are ad placements, course enrollment systems, and enhanced user interface automation.

**Acknowledgements:** This research is supported by the HP Labs Innovation Research Program.